\documentclass[screen]{acmart}

\usepackage{booktabs} %

\usepackage{graphicx}  %
\usepackage{amsfonts} 
\usepackage{CJK}
\usepackage{bm}
\usepackage{multirow}
\usepackage{subfigure}
\usepackage{makecell}
\usepackage{xspace}
\usepackage{array}
\usepackage{colortbl}
\usepackage{xcolor}

\newcommand{\ie}{\emph{i.e.,}\xspace}

\newcommand{\eg}{\emph{e.g.,}\xspace}
\newcommand{\etal}{\emph{et al.}\xspace}
\newcommand{\ignore}[1]{}

\newcommand{\dubbelop}{$^{\blacktriangle}$}

\newcommand{\dubbelneer}{$^{\blacktriangledown}$}

\AtBeginDocument{%
  \providecommand\BibTeX{{%
    \normalfont B\kern-0.5em{\scshape i\kern-0.25em b}\kern-0.8em\TeX}}}

\setcopyright{acmcopyright}
\acmJournal{TOIS}
\acmYear{2020} \acmVolume{1} \acmNumber{1} \acmArticle{1} \acmMonth{1} \acmPrice{15.00}

\title{Meaningful Answer Generation of E-Commerce Question-Answering}

\begin{document}

\author{Shen Gao}
\authornote{Equal contribution. Ordering is decided by a coin flip.}
\affiliation{%
  \institution{Wangxuan Institute of Computer Technology, Peking University}
}
\email{shengao@pku.edu.cn}

\author{Xiuying Chen}
\authornotemark[1]
\affiliation{%
  \institution{Wangxuan Institute of Computer Technology, Peking University}
}
\email{xy-chen@pku.edu.cn}

\author{Zhaochun Ren}
\affiliation{%
  \institution{School of Computer Science and Technology, Shandong University}
}
\email{zhaochun.ren@sdu.edu.cn}

\author{Dongyan Zhao}
\affiliation{%
    \institution{Wangxuan Institute of Computer Technology, Peking University}
}
\email{zhaody@pku.edu.cn}

\author{Rui Yan}
\authornote{Corresponding Author: Rui Yan (ruiyan@pku.edu.cn)}
\affiliation{
  \institution{\textsuperscript{1} Gaoling School of Artificial Intelligence, Renmin University of China; 
  \textsuperscript{2} Wangxuan Institute of Computer Technology, Peking University}
}
\email{ruiyan@pku.edu.cn}

\begin{abstract}
In e-commerce portals, generating answers for product-related questions has become a crucial task. 
In this paper, we focus on the task of \emph{product-aware answer generation}, which learns to generate an accurate and complete answer from large-scale unlabeled e-commerce reviews and product attributes.

However, \textit{safe answer problems} (\ie neural models tend to generate meaningless and universal answers) pose significant challenges to text generation tasks, and e-commerce question-answering task is no exception.
To generate more meaningful answers, in this paper, we propose a novel generative neural model, called the \emph{Meaningful Product Answer Generator} (MPAG), which alleviates the safe answer problem by taking product reviews, product attributes, and a prototype answer into consideration.
Product reviews and product attributes are used to provide meaningful content, while the prototype answer can yield a more diverse answer pattern.
To this end, we propose a novel answer generator with a review reasoning module and a prototype answer reader.
Our key idea is to obtain the correct question-aware information from a large scale collection of reviews and learn how to write a coherent and meaningful answer from an existing prototype answer.
To be more specific, we propose a read-and-write memory consisting of selective writing units to conduct \textit{reasoning among these reviews}.
We then employ a prototype reader consisting of comprehensive matching to extract the \textit{answer skeleton} from the prototype answer.
Finally, we propose an answer editor to generate the final answer by taking the question and the above parts as input.
Conducted on a real-world dataset collected from an e-commerce platform, extensive experimental results show that our model achieves state-of-the-art performance in terms of both automatic metrics and human evaluations. %
Human evaluation also demonstrates that our model can consistently generate specific and proper answers.

\end{abstract}

\ccsdesc[300]{Information systems~Question answering}

\keywords{Question-answering, e-commerce, product-aware answer generation}

\maketitle

\section{Introduction}
\label{sec:intro}

In recent years, the explosive popularity of \emph{question-answering} (QA) is revitalizing the task of \emph{reading comprehension} with promising results~\cite{wang2017gated,Song2017Summarizing}.
Unlike traditional knowledge-based QA methods that require a structured knowledge graph as the input and output resource description framework (RDF) triples~\cite{He2017GeneratingNA}, most of the reading comprehension approaches read context passages and extract text spans from input text as answers~\cite{Seo2017bi, wang2017gated}.

\begin{figure}
    \centering
    \includegraphics[scale=0.75]{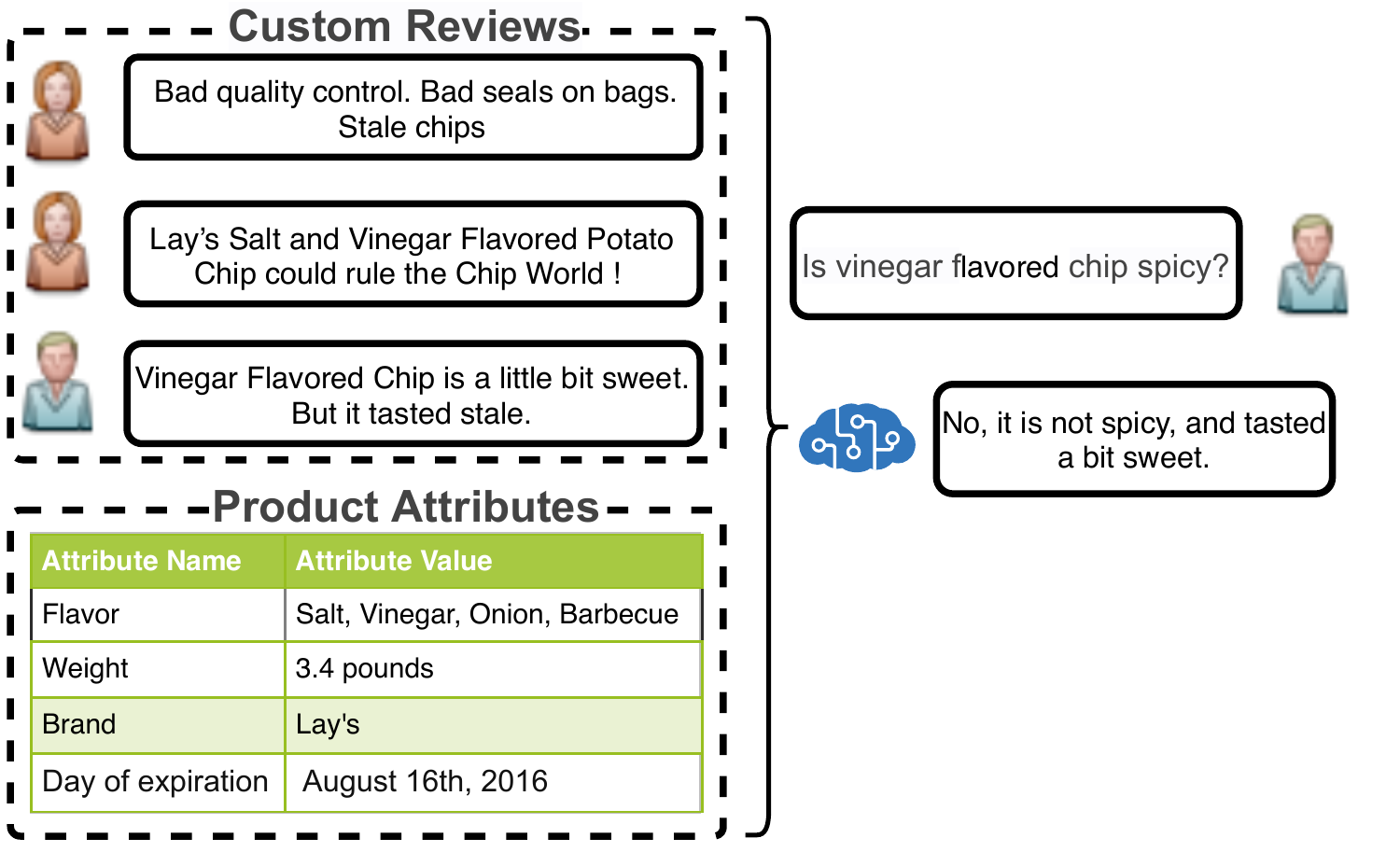}
    \caption{
    Example of giving an answer for product related question by sourcing from custom reviews and product attributes.
    }
    \label{fig:intro-case-1}
\end{figure}

E-commerce is playing an increasingly important role in our daily life.
As a convenience of users, more and more e-commerce portals provide community question-answering services that allow users to pose product-aware questions to other consumers who purchased the same product before.
Unfortunately, many product-aware questions lack proper answers.
Under the circumstances, users have to read the product's reviews to find the answer by themselves. 
Given product attributes and reviews, an answer is manually generated following a cascade procedure:
\begin{enumerate}
    \item A user skims reviews and finds relevant sentences;
    \item She/he extracts useful semantic units;
    \item The user jointly combines these semantic units with attributes, and writes a proper answer.
\end{enumerate}
However, the information overload phenomenon makes this procedure an energy-draining process to pursue an answer from a rapidly increasing number of reviews.
Consequently, automatic product-aware question-answering has become more and more helpful in this scenario.
In this paper, the task on which we focus is the \emph{product-aware answer generation}.
Our goal is to respond product-aware questions automatically given a large number of reviews and attributes of a specific product.
Figure~\ref{fig:intro-case-1} shows an example of product-aware answer generation task.
Unlike either a ``yes/no'' binary classification task~\cite{McAuley2016Addressing} or a review ranking task~\cite{Moghaddam2011AQAAO}, product-aware answer generation provides a natural-sounding sentence as an answer.

The definition of our task is similar to the reading comprehension~\cite{wei2018fast,Rajpurkar2016SQuAD10} which reads some paragraphs and then answers the question by extracting text spans as the response. 
The knowledge source of the reading comprehension task always comes from formal documents, like news articles and Wikipedia.
However, product reviews from e-commerce websites are informal and noisy, whereas in reading comprehension the given context passages are usually in a formal style. 
Generally, using existing reading comprehension approaches to tackle the product-aware answer generation confronts three challenges:
\begin{enumerate}
    \item Some of the review texts are irrelevant and noisy;
    \item It's extremely expensive to label large amounts of explicit text spans from real-world e-commerce platforms;
    \item Traditional loss function calculation in reading comprehension tends to generate meaningless answers such as ``I don't know''.
\end{enumerate}

To overcome these drawbacks, we propose the \textit{product-aware answer generator} (PAAG) in our early work~\cite{Gao2019ProductAware}, a product related question answering model which incorporates customer reviews with product attributes. 
Specifically, at the beginning, we employ an attention mechanism to model interactions between a question and reviews.
Simultaneously, we employ a key-value memory network to store the product attributes and extract the relevance values according to the question.
Eventually, we propose a recurrent neural network (RNN) based decoder, which combines product-aware review representation and attributes to generate the answer.
More importantly, to tackle the problem of meaningless answers, we propose an adversarial learning mechanism for optimizing parameters.
To demonstrate the effectiveness of our proposed model, we collect a large-scale e-commerce question answering dataset from one of the largest online shopping websites JD.com.
Experimental results conducted on our proposed dataset demonstrate that the PAAG model achieves significant improvement over other baselines, including the state-of-the-art reading comprehension model.
Our experiments verify that adversarial learning is capable to significantly improve the denoising and facts extracting capacity of PAAG.

\begin{figure}
    \centering
    \includegraphics[width=\columnwidth]{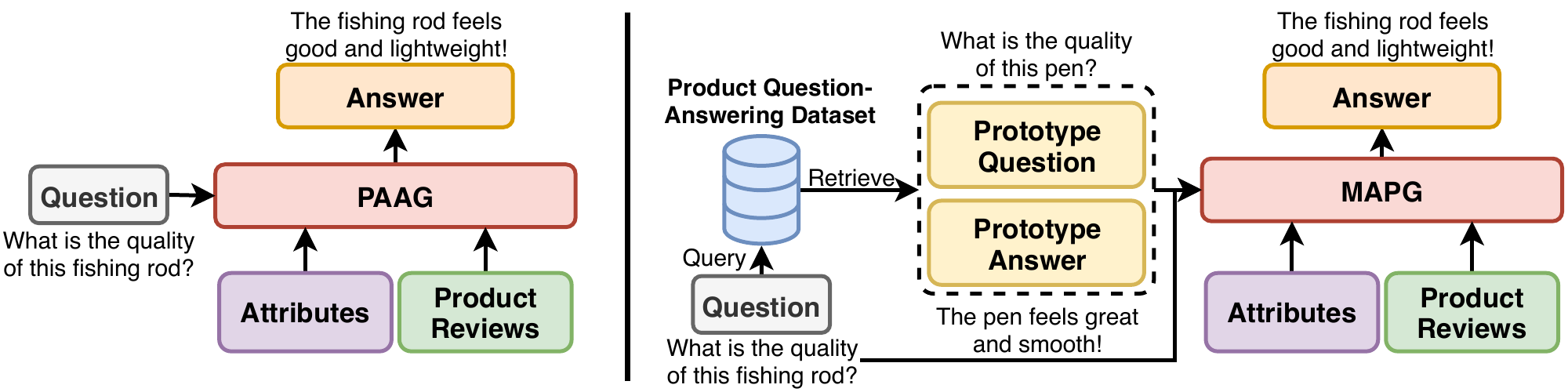}
    \caption{
    Illustration the data flow of two product-related question answering model. The PAAG model (left) only use the question, product attributes and custom reviews as input. To overcome the safe answer problem, MAPG (right) incorporates the retrieved prototype question and answer pair additionally, which can give a natural answer pattern and boost the performance of answer generation.
    }
    \label{fig:intro-case}
\end{figure}

Although using adversarial learning techniques can reduce the probability of generating safe answers to a certain extent, the model still has a high probability of generating safe answers since the model lacks answer patterns that can be referenced.
This is known as the \emph{safe answer problem}, which is a commonly-faced problem in text generation tasks~\cite{Li2016ADO,xing2017topic}.
Moreover, to answer some complex questions, the QA system needs the reasoning ability that could make inference from the product reviews.

Motivated by this observation, in this work, we take one step further and improve our previously proposed PAAG framework that addresses the safe answer problem in e-commerce question-answering.
To be more specific, we solve the problem by introducing a large-scale collection of reviews and a prototype question-answer pair and employing the memory network to incorporate the reasoning result into the answer generation process.
Existing works~\cite{Gao2019Product,Chen2019Review} only employ a limited number of reviews (less than 10) and are thus inclined to generate a biased and inaccurate answer.
To avoid this, we target at learning accurate and appropriate content information from massive amounts of product reviews and product attributes.
However, discovering and utilizing information from large quantities of reviews is highly challenging.
Sometimes the answer generation model needs to do reasoning among the reviews to obtain the final facts that are necessary for generation answers.
Then, to learn a more diverse and interesting answer pattern, the prototype answer can be of great help.
The prototype answer gives a natural language pattern of a similar question, and it can be referred by the answer generation model.
In fact, existing approaches~\cite{Wu2018ResponseGB,Guu2018GeneratingSB} in the dialog generation field have proven the usefulness of incorporating prototype response for improving performance, which retrieves a prototype text and then post-edit it as the final dialog response.
Nevertheless, combining prototype text and reasoning results with the question-answering task has yet to be explored.

In this paper, we propose a novel answer generation model, named \emph{Meaningful Product Answer Generator} (MPAG), for e-commerce question answering.
Figure~\ref{fig:intro-case} illustrates the framework of MPAG, which takes the question and three additional information sources as input: product reviews, product attributes, and a prototype question with an answer.
First, for product reviews, MPAG uses a simple but efficient clustering method, K-means, to aggregate similar reviews to the same cluster, so as to better utilize review information.
Then, we employ Convolutional Neural Network (CNN) to encode these reviews.
To reason about these reviews, we propose a write-read memory architecture that selectively writes review information to the memory, and then reads out corresponding information related to the question.
Next, MPAG employs a key-value memory network to encode product attributes.
To tackle the safe answer problem, we retrieve a prototype answer from the dataset and employ a prototype reader to learn the answer skeleton.
Specifically, we use the question to retrieve the most similar question from the dataset as the prototype question and use the answer to this prototype question as prototype answer.
Finally, we propose an answer editor to incorporate the answer skeleton with the reasoning result and product attributes, and then generate the new answer.
Experiments conducted on a public large-scale benchmark dataset demonstrate that MPAG achieves significant improvement over the state-of-the-art baselines. 
Experiments also verify the effectiveness of each module in MPAG as well as its explanation ability.

This work is a substantial extension of our previous work reported at WSDM 2019~\cite{Gao2019Product}. 
The extension in this article includes a novel memory network and a prototype editing-based answer generator, a proposal of a new framework for answering the product-related questions in e-commerce portals which can generate more meaningful answers than the previous method. 
Specifically, the contributions of this work include the following:

\begin{itemize}
    \item We come up with a meaningful answer generator model in the e-commerce question-answering task.
    \item We propose a review reasoning module to reason about a large number of reviews.
    \item  We employ a prototype editing based answer generator to generate answers by revising a given prototype answer and incorporating the reasoning results.
    \item Experiments conducted on a public large-scale benchmark dataset show that our model outperforms all baselines, including state-of-the-art models.
    Experiments also verify the effectiveness of each module in MPAG, as well as its interpretability in answer generation.
\end{itemize}

The rest of the paper is organized as follows: 
We introduce related work in \S~\ref{section2}. 
We formulate our research problem in \S~\ref{sec:formulation}.
We introduce our extended method which incorporates answer prototype and a novel memory network in \S~\ref{mpag-model}. 
Then, \S~\ref{section5} details our experimental setup and \S~\ref{section6} presents the experimental results. 
Finally, \S~\ref{section7} concludes the paper. 
\section{Related Work}
\label{section2}

In this section, we detail related work on product-aware question-answering, reasoning in question-answering, reading comprehension, text generation methods and prototype editing, respectively.

\subsection{Product-aware Question-answering} 
Studies on product reviews include~\cite{wu2017reviewminer,hyun2018review,wang2018sentiment}. 
It is a long-standing issue in information retrieval.
Most of the existing strategies for product-aware question-answering aim at extracting relevant sentences from input reviews to answer the given question~\cite{yu2012answering,yu2018aware,McAuley2016Addressing}.
For example, \citet{yu2012answering} proposed an opinion-based question-answering framework, which organizes reviews into a hierarchical structure and retrieves a review sentence as the answer. 
With the development of knowledge graphs, reviews have been considered as external knowledge~\cite{McAuley2016Addressing} to predict the answer.
However, it can only return simple answers, such as ``yes'' or ``no''.
Unfortunately, more often than not there is no proper review that can be used as an answer.
\citet{Gupta2019AmazonQAAR} proposed a review-based question answering dataset.
However, with this dataset, a high BLEU score (with a 78.56 BLEU-1 score) can be achieved by just randomly selecting a review as the answer.
Thus, it is not suitable for generative question answering tasks, which have gained particular interest in recent years due to the emergence of neural networks.
For example, \citet{Gao2019Product} proposed an adversarial learning-based model which combines product attributes and review information to generate an answer for a given question.
\citet{Chen2019Review} proposed a convolutional text generation model which uses review snippets to guide the decoding attention.
However, these generation-based approaches are not sufficiently robust against the safe answer problem, and they also lack reasoning ability.

\subsection{Reasoning in Question-answering} 
Question-answering has long been a task used to assess a model's ability to understand and reason about language. 
Large scale datasets such SQuAD~\cite{rajpurkar2016squad} have encouraged the development of many advanced, high performing attention-based neural models.
The ability of reasoning is an important research ingredient in question-answering~\cite{weston2015towards,Sukhbaatar2015EndToEndMN,Tay2019SimpleAE}.
\citet{weston2015towards} released a dataset, named bAbI, to specifically focus on multi-step reasoning by requiring models to reason using disjoint pieces of information.
For this task, iteratively updating the query representation with context information has also been shown to effectively emulate multi-step reasoning.
\citet{kumar2016ask} proposed a dynamic memory network where questions trigger an iterative attention process to condition the model's attention on inputs and the result of previous iterations.
\citet{xiong2016dynamic} proposed several improvements to memory and input modules and apply them to visual question-answering.
Instead of extractive fact-finding QA, \citet{bauer2018commonsense} focused on a multi-hop generative task, which requires the model to reason, gather, and synthesize disjoint pieces of information within the context to generate an answer.
Apart from multi-hop based reasoning, reasoning with a memory write-read mechanism has also been considered~\cite{graves2016hybrid,le2019learning,csordas2018improving,Wang2018M3}.
For instance, \citet{graves2016hybrid} proposed a machine learning model, namely differentiable neural computer, which consists of a neural network that can read from and write to an external memory matrix, analogous to the random-access memory in a conventional computer.
Subsequently, \citet{le2019learning} proposed a modified version aiming to balance between maximizing memorization and forgetting via overwriting mechanisms.

\smallskip \noindent
\noindent Inspired by the above solutions, we also apply the write-read mechanism in this paper. 
In contrast with DNC, we come up with a novel reasoning module, which is simpler and more efficient for e-commerce question-answering.

\subsection{Reading Comprehension}
Given a question and relevant passages, reading comprehension extracts a text span from passages as an answer~\cite{Rajpurkar2016SQuAD10}. Recently, based on a widely applied dataset, i.e., SQuAD~\cite{Rajpurkar2016SQuAD10}, many approaches have been proposed~\cite{Sun2019Improving,Cao2020Unsupervised,Dua2020Benefits,Hu2019Read,Mihaylov2019DiscourseAware}.
Seo \etal~\cite{Seo2017bi} use bi-directional attention flow mechanism to obtain a query-aware passage representation.
Wang \etal~\cite{wang2017gated} propose a model to match the question with passage using gated attention-based recurrent networks to obtain the question-aware passage representation.
Consisting exclusively of convolution and self-attention, QANet~\cite{wei2018fast} achieves the state-of-the-art performance in reading comprehension.
\citet{Cui2017Attention} place another attention mechanism over the document-level attention and induces ``attended attention'' for final answer predictions.
As mentioned above, most of the effective methods contain question-aware passage representation for generating a better answer.
This mechanism makes the models focus on the important part of passage according to the question.
Following these previous studies, our method models the reviews of product with a question aware mechanism.

\subsection{Text Generation Methods} 

In recent years, sequence-to-sequence (seq2seq)~\cite{Sutskever2014SequenceTS} based neural networks have been proved effective in generating a fluent sentence.
The seq2seq model is originally proposed for machine translation and later adapted to various natural language generation tasks, such as text summarization~\cite{Wang2019Concept,Paulus2018A,Gehrmann2018BottomUp,Lin2018Global,Wang2019BiSET,Chen2019Learning,Gao2020From,Gao2019Abstractive,Gao2019Learning} and dialogue generation~\cite{Tao2018RUBERAU,Yao2017TowardsIC,Cai2019Retrievalguided,Zhao2020LowResource,Zhang2020Modeling,Gao2020Learning,Gao2020Preference,qiu2020what,fu2020query,Li2019InsufficientDC}.
Rush \etal~\cite{Rush2015ANA} apply the seq2seq mechanism with attention model to text summarization field.
Then See \etal~\cite{See2017GetTT} add copy mechanism and coverage loss to generate summarization without out-of-vocabulary and redundancy words.
Tao \etal~\cite{Tao2018Get} propose a multi-head attention mechanism to capture multiple semantic aspects of the query and generate a more informative response.
Yao \etal~\cite{Yao2017TowardsIC} propose to use the content introducing method to solve the problem of generating meaningless response.
Wang \etal~\cite{Wang2018Chat} use three channels for widening and deepening the topics of interest and try to make the conversational model chat more turns.

Different from vanilla seq2seq models, our model utilizes not only the information in input sequence but also much external knowledge from user reviews and product attributes to generate the answer that matches the facts.
Similar to our e-commerce question answering task, several tasks input data in key-value structure instead of a sequence.
In order to utilize these data when generating text, key-value memory network (KVMN)~\cite{Abdelrahman2019Knowledge,Xu2019Enhancing} is purposed to store this type of data.
He \etal~\cite{He2017GeneratingNA} incorporate copying and retrieving knowledge from knowledge base stored in KVMN to generate natural answers within an encoder-decoder framework. 
In detail, they retrieve some relative facts and store them in a KVMN fashion, then use an attention mechanism to attend the facts and fuse them into context vector.
Tu \etal~\cite{Tu2018Learning} use a KVMN to store the translation history which gives model the opportunity to take advantage of document-level information instead of translate sentences in an isolation way.
In view of the superior performance of storing structure data in neural models, we employ the key-value memory network in our model to store the attributes data of product.

\subsection{Prototype Editing} 

The safe answer problem has been widely explored in recent years~\cite{Li2015ADO,qiu-etal-2019-training,hashimoto2018edit}.
Among these methods that aim at solving this challenge, prototype editing has proven one of the most effective.
\citet{Guu2018GeneratingSB} were the first to propose the prototype editing model, where a prototype sentence is sampled from the training data and edited into a new sentence.
Subsequently, \citet{Wu2018ResponseGB} proposed a new paradigm for response generation, which first retrieves a prototype response from a pre-defined index and then edits it according to the differences between the prototype context and current context.
Different from this soft attention method, \citet{cai2018skeleton} proposed a hard-editing skeleton-based model to promote the coherence of the generated stories.
Specifically, a skeleton is generated by revising the retrieved responses; then, a generative model uses both the generated skeleton and the original query to generate a response.
\citet{Cao2018Retrieve} applied this prototype editing method to the task of summarization, where they employed existing summaries as soft templates to generate a summary.

While previous prototype-based methods have achieved much success in various areas, none have incorporated reviews or product attributes into their generation, limiting their ability to produce appropriate and accurate answers.
Thus, our proposed method is the first attempt to apply the prototype editing method to question-answering, taking advantage of reasoning results and product attributes to generate an answer.
The differences of technical design between our model and previous prototype-based methods lie in that these methods directly use the attention mechanism~\cite{Wu2018ResponseGB,Cao2018Retrieve} to obtain the edit vector, which ignore the relationships between the prototype answer and prototype question.
And this relationship can help our model to identify which part in the prototype has a low correlation with the prototype question, and that part will be used as the answer prototype.

\section{Problem formulation}
\label{sec:formulation}

Before detailing our answer generation model, we first introduce our notations listed in Table~\ref{tbl:notations}.

\begin{table}[!t]
 \caption{Glossary.}
 \label{tbl:notations}
 \centering
 \begin{tabular}{ll}
  \toprule
  Symbol & Description \\
  \midrule 
  $X^q$ & a question sentence \\
  $X^r$ & a set of reviews sentences \\
  $A$ & a set of product attribute key-value pairs  \\
  $a^k_t, a^v_t$ & the $t$-th attribute key and value in $A$\\
  $Y, \hat{Y}, \Tilde{Y}$ & ground truth answer, generated answer, prototype answer \\
  $x^r_{n,t}$ & $t$-th word in $n$-th review \\
  $x^q_t, \hat{y}_t, \Tilde{y}_t$ & $t$-th word in corresponding sentence \\
  $N$ & number of reviews in a cluster \\
  $K$ & number of clusters \\
  $C_k$ & $k$-th review cluster \\
  $T^n_r$ & length of $n$-th review \\
  $T_a$ & number of attributes \\
  $T_q, T_y$ & length of corresponding word sequence \\
  \bottomrule
 \end{tabular}
\end{table}

For a product, there is a question $X^q = \{x^q_1, x^q_2, \dots,\allowbreak x^q_{T_q}\}$, along with reviews $X^r = \{x^r_1, x^r_2, \dots, x^r_{T_r}\}$, where $T_{r}$ represents the number of reviews, $x_{t}^q$ is the $t$-th word in question and answer and $x_t^r$ is the $t$-th review. 
We assume there exist $T_{a}$ key-value pairs of product attributes $A = \{(a^k_1, a^v_1), (a^k_2, a^v_2), \dots, (a^k_{T_a}, a^v_{T_a})\}$, where $a^k_i$ is the name of $i$-th attribute and $a^v_i$ is the attribute content.
Both key $a^k_i$ and value $a^v_i$ include one word.
Since our newly proposed model is a prototype-based method, a prototype question $\Tilde{X}^q = \{\Tilde{x}^q_1, \Tilde{x}^q_2, \dots, \Tilde{x}^q_{T_q}\}$ and prototype answer $\Tilde{Y} = \{\Tilde{y}_1,\allowbreak \Tilde{y}_2, \dots, \Tilde{y}_{T_y}\}$, where $T_q$ and $T_y$ is the number of words in prototype question and answer, is also attached.
The goal of our task is to generate an answer $\hat{Y}$ that is in accordance with product attributes $A$ and information mentioned in reviews $X^r$.

We formulate the Meaningful Product Answer Generator (MPAG) as follows: 
Given a question $X^q$, MPAG first reads reviews $X^r$ and attributes $A$, then generates an answer $\hat{Y} = \{\hat{y}_1, \hat{y}_2, \dots, \hat{y}_{T_y}\}$ via editing the prototype answer $\Tilde{Y}$.
That is, the generator maximizes the probability $P(Y|X^q, X^r, A) = \prod_{t=1}^{T_y} P(y_t|X^q, X^r, A,\Tilde{X}^q, \Tilde{Y})$, where $Y = \{y_1, y_2, \dots, y_{T_y}\}$ is the ground truth answer.

\section{MPAG Model}
\label{mpag-model}

Although our previous proposed PAAG model employs an adversarial learning strategy to encourage the model to generate meaningful answers, and that training method punishes the model when generating answers which do not include the correct product facts, the PAAG model still tends to generate safe answers, like ``ask the custom service'' or ``I don't know''.
In this paper, we propose to explicitly introduce a natural answer pattern to the answer generation model, which can be used when generating new answers.
Moreover, to extract accurate information from the reviews to form the answer, we propose a novel memory architecture to reason from the reviews.
Then, we jointly incorporate the answer pattern and the reasoning results in the final answer generation process.
We argue that these extensions will increase the performance of generating more accurate answers for product related questions.

\begin{table}[t]
    \centering
    \caption{Comparision between PAAG and MAPG.}
    \begin{tabular}{@{}lcc@{}}
    \toprule
    & PAAG & MAPG \\ 
    \midrule
    Question Encoder & RNN & RNN \\
    Review Encoder & RNN & SRU \\
    Review Clustering & - & K-means \\
    Review Reasoning & - & Read-Write Memory \\
    Product Attribute Encoding & Key-Value Memory & Key-Value Memory \\
    Propotype Reader & - & Answer Skeleton Extractor \\
    Decoder & RNN & Editing Gating \\
    Training Stratege & Adversarial Training & NLL \\
    \bottomrule
    \end{tabular}
    \label{tab:comp_model}
    \end{table}

Although we use the same attribute encoder (key-value based memory network) and question encoder (RNN-based text encoder) with the PAAG model, there are three significant differences in our MPAG model compared with PAAG:
\begin{enumerate}
    \item The PAAG only leverages a few reviews and uses a simple attention based interaction module with the question. And in this paper, our model uses many reviews as input and employs a clustering method and reasoning module to extract useful information from these reviews.
    \item We incorporate prototype question-answer pair to facilitate the answer generation process, which is not used in PAAG.
    \item We propose to use the editing gate to fuse the information from answer skeleton and reasoning result dynamically when generating the answer.
\end{enumerate}
Specifically, we show the comparison between PAAG and MAPG in Table~\ref{tab:comp_model}.

\subsection{Overview}

\begin{figure*}
    \centering
    \includegraphics[scale=0.7]{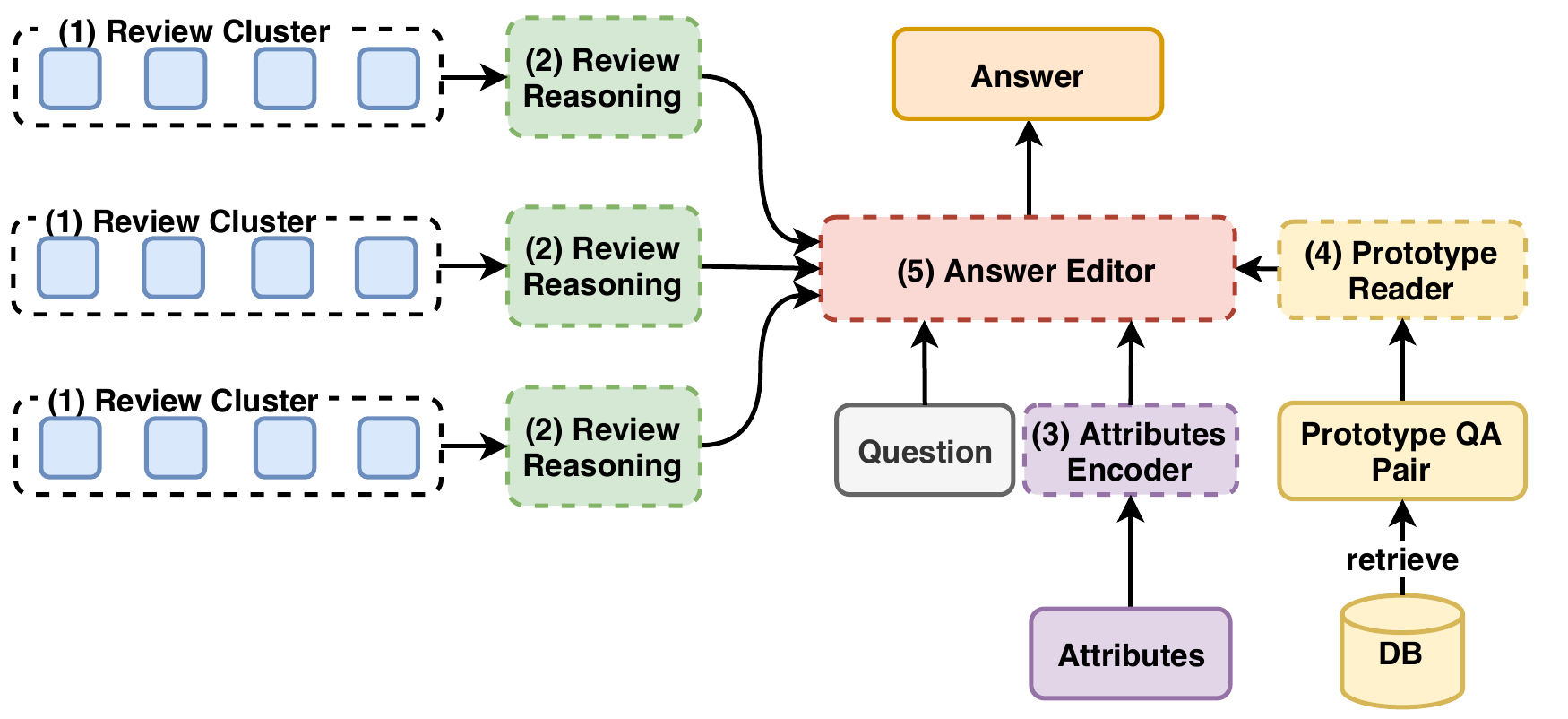}
    \caption{
    Overview of MPAG. We divide our model into five ingredients: (1) \textit{Review clustering module} aggregates the review into $K$ clusters by using the K-means clustering method on the bag-of-word (BOW) vector of all the reviews; (2) \textit{Review reasoning module} uses a read-write memory to reason over the reviews in each cluster and produce the reasoning result; 
    (3) \textit{Attributes encoder} learns a question-aware attribute representation; (4) \textit{Prototype reader} generates an answer skeleton from prototype answer to enhance the diversity of sentence pattern; (5) \textit{Answer editor} fuses the result from previous stages by an editing gate and generates an answer.
    }
    \label{fig:overview}
\end{figure*}

In this section, we introduce our meaningful product answer generator model in detail.
The overview of MPAG is shown in Figure~\ref{fig:overview} and can be split into five modules:
(1) Review clustering (See \S\ref{sec:clustering}): We employ the K-means algorithm to aggregate reviews into clusters.
(2) Review reasoning module (See \S\ref{sec:memory}): We employ a write-read memory reasoning framework to reason among all reviews to learn useful information from the reviews according to the question.
(3) Attributes encoder (See \S\ref{sec:attributes}): We encode and extract attributes related to the question by key-value memory network.
(4) Prototype reader (See \S\ref{sec:prototype}): We use a recurrent network to model the prototype context and extract prototype answer skeleton that can be reused.
(5) Answer editor (See \S\ref{sec:editor}): Eventually, we employ an RNN-based decoder to generate the answer incorporating prototype skeleton, reasoning result, and attribute representation.

\subsection{Review clustering}
\label{sec:clustering}
Since our model takes a large number of reviews focusing on different aspects as input, processing them together will confuse the model and make it hard to learn useful information from them.
Thus, we employ a clustering step to aggregate similar reviews into the same cluster.
To begin with, we use the bag-of-word (BOW) vector to represent each review sentence.
We then employ the K-means algorithm to aggregate these reviews into $K$ clusters.
In each cluster, if the number of reviews is less than $N$, we append empty review to the cluster to pad the cluster.
Conversely, if the number of reviews is larger than $N$, we drop some review to keep each cluster to contain exactly $N$ reviews.
These clustering reviews can be denoted as $X^r = \{C_1, \dots, C_K\}$, where $C_k$  denotes the $k$-th review cluster which contains $N$ reviews.

\subsection{Review reasoning module}
\label{sec:memory}

\begin{figure*}
    \centering
    \includegraphics[scale=0.80]{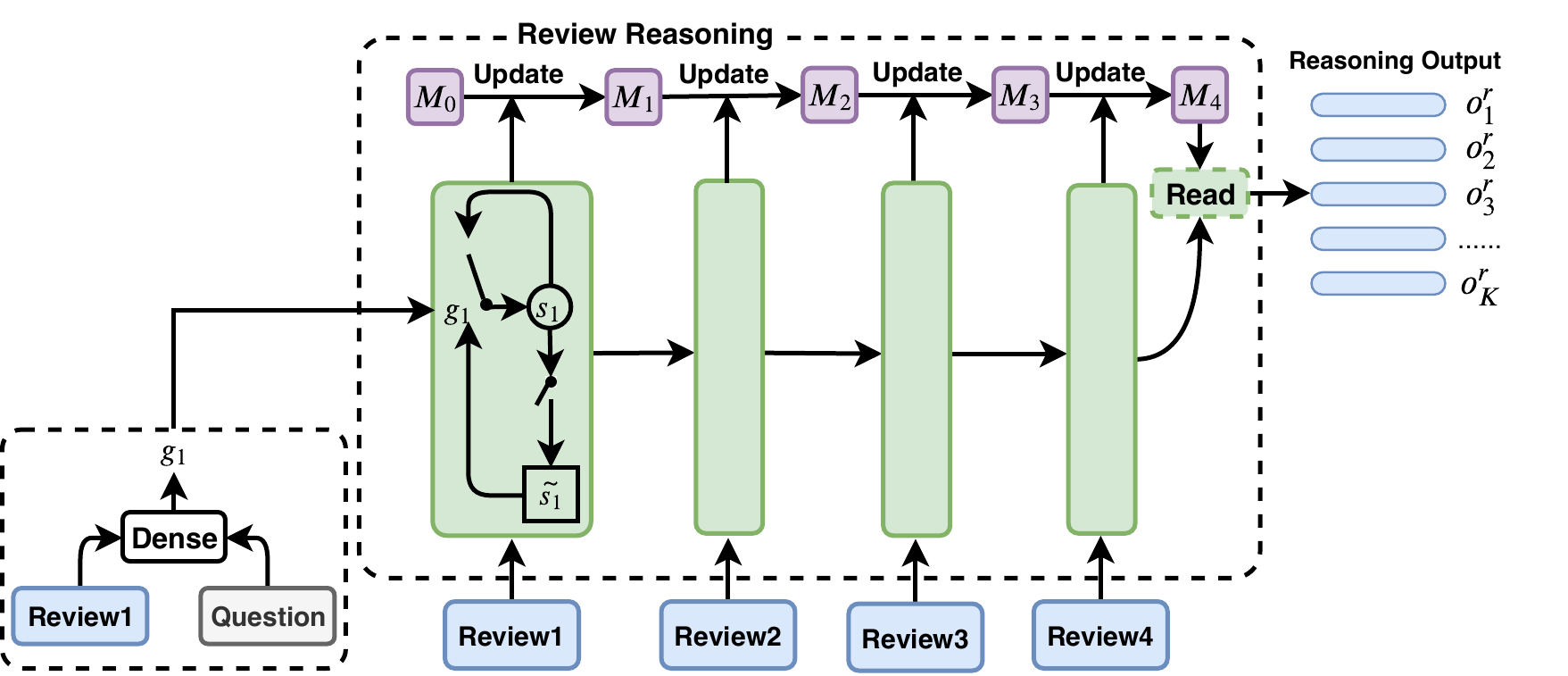}
    \caption{
    Overview of the review reasoning module in MPAG. In the beginning, the selective recurrent unit is applied on each review; then retrieved information from each review is written to memory; finally, a multi-head memory reading mechanism is employed to read the memory.
    }
    \label{fig:memory}
\end{figure*}

For each review cluster, we employ a reasoning module to conduct reasoning among the reviews in each cluster separately.
Figure~\ref{fig:memory} illustrates the whole process. 
In this section, we omit the subscript $k$ of cluster index for simplicity.

For question $X^q$, we use an embedding function $e$ to map one-hot representation of each word $x^q_{t} \in X^q$ into a high-dimensional vector space, $e(x^q_{t})$.
(The words in reviews $X^r$, attributes $A$, prototype question $\Tilde{X}^q$ and answer $\Tilde{Y}$ are also embedded in this way.)
We then employ a \emph{bi-directional recurrent neural network} (Bi-RNN) to model the temporal interactions between words in $X^q$, so we have:
\begin{align}
    h^q_t &= \text{Bi-RNN}_q(e(x^q_t) h^q_{t-1}), \label{eq:question-encode}
\end{align}
\noindent where $h^{q}_t$ denotes the hidden state of $t$-th step in Bi-RNN for question $X^q$. 
We use the final hidden state $h^q_{T_q}$ of $\text{Bi-RNN}_q$ to represent the whole question sentence $X^q$.
We here choose the \emph{long short-term memory} (LSTM) as the cell of Bi-RNN. 
One can also replace LSTM with similar algorithms such as \emph{Gated Recurrent Unit} (GRU)~\cite{Chung2014EmpiricalEO}.
We leave the study for future work.

To extract the semantic features from each review, we first employ an CNN with a max-pooling operation, then apply a \emph{Selective Reading Unit} (SRU)-based RNN to obtain final representation for each review.
To begin with, a list of kernels with different width are used in the CNN operation, and their outputs are concatenated together, denoted as $h^r_{n,t}$ in Equation~\ref{eq:review-cnn}.
These different kernels capture different n-grams features.
A max-pooling operation is then conducted to extract the most salient feature from the output of CNN, shown in Equation~\ref{eq:review-maxpool}:
\begin{align}
    h^r_{n,t} &= \text{CNN} \left( e(x^r_{n,t}) \right), \label{eq:review-cnn}\\
    h^r_{n} &= \text{max-pool} (\{h^r_{n,1}, \dots, h^r_{n,T^{n}_r}\}), \label{eq:review-maxpool}
\end{align}
where $x^r_{n,t}$ denotes the $t$-th word in $n$-th review, $T^{n}_r$ is the length of $n$-th review, and $h^r_{n}$ is the vector representation of $n$-th review.  %

Since we need to identify the salient review to give the answer of current question, the relationship between review and question should be considered when generating the representation for each review.
Inspired by GRU~\cite{Chung2014EmpiricalEO}, to further study the interactions between reviews, we establish an RNN made up of SRUs.
First, in Equation~\ref{eq:sru-n}, we fuse the representation of question and review together.
Then we conduct a dense layer on the fusion representation $n_i$ (see Equation~\ref{eq:sru-z}) and normalize the update gate $g_i$ over $N$ steps using softmax function (see Equation~\ref{eq:salience-score}).
For $i$-th step (review), SRU calculates an update gate $g_{i}$, which is decided by question and review together, as shown in Equation~\ref{eq:salience-score}:
\begin{align}
n_{i} &= [h^r_{i} \times h^q_{T_q}; h^r_{i}; h^q_{T_q}], \label{eq:sru-n}\\
z_{i} &= W_{2}\tanh(W_{1}n_{i}+b_{1})+b_{2},  \label{eq:sru-z}\\
g_{i} &= \frac{\exp(z_{i})}{\sum^N_{j=1} \exp(z_{j})}, \label{eq:salience-score}
\end{align}
where $\times$ denotes the element-wise multiplication.
Note that the review with a high $g_{i}$ value means that the information of this review should be mostly retained, and will play a more important role when generating the answer.
Thus, the update gate $g_{i}$ can also be seen as the \textbf{salience weight} of $i$-th review.

Unlike GRU, SRU incorporates the question representation $h_{T_{q}}^{q}$ into the calculation of update gate which can help the model to identify which review has more contribute on answering current question.
Then update gate $g_i$ is used in updating the hidden state $s_{i}^r$, shown in Equation~\ref{eq:gru-hi}:
\begin{align}
q_{i} &= \sigma(W_{q}h^r_{i}+U_{q}s_{i-1}^r+b_{q}),  \\
\tilde{s_{i}^r} &= \text{tanh}(W_{s}h^r_{i} + q_{i} \times U_{s} s_{i-1}^r + b_{s}) ,\\
s_{i}^r &= g_{i} \times \tilde{s_{i}^r} + (1-g_{i}) \times s_{i-1}^r.\label{eq:gru-hi}
\end{align}
We use the hidden state of $i$-th step $s_{i}^r$ asthe final representation of $i$-th review.

Now we have a more comprehensive representation $s_{i}^r$ for each review. 
Next, we focus on conducting reasoning among these reviews, and we propose a review reasoning memory network with a write-read mechanism~\cite{graves2016hybrid}.
As preparation, we initialize an empty memory matrix $M_0 \in \mathbb{R}^{S \times H}$ with $S$ memory slots.
Each slot is a $H$-dimension vector which is set as a representation of learned information and reviews with the same information will be written into the same slot. 
We use the notion $m^0_j$ to denotes the $j$-th slot in memory $M_0$.

\subsubsection{Writing to memory}

In this section, we describe the memory writing process that writes each review representation into memory one by one, from $s_1^r$ to $s_N^r$, and the memory is updated from $M_0$ to $M_N$ simultaneously.
In each step, MPAG reads a review representation $s_i^r$ and writes it to the memory matrix $M_i$.
To update the memory matrix, this module calculates $3$ components: write weight for each slot, a write content vector, and an erase vector.

We now detail our writing process. 
To begin with, we calculate a \emph{write weight} for $i$-th review, ensuring that similar reviews will have similar write weights. 
The module first calculates a write key $\kappa^w_i \in \mathbb{R}^H$ using a dense layer applied on the input review representation $s_{i}^r$:
\begin{equation}
    \kappa^w_i = \text{Dense}(s_{i}^r),
\end{equation}
where the write key $\kappa^w_i$ contains the information learned from the $i$-th review.
We obtain the \emph{write weight} $\pi^w_{i,j} \in \mathbb{R}$ for the $j$-th memory slot $m^i_j$ by calculating the similarity between $\kappa^w_i$ and memory slot $m^i_j$:
\begin{align}
    \pi^w_{i,j} &= \frac{\exp(\mathcal{S}(\kappa^w_i, m^i_j))}{\sum^S_{l=1} \exp(\mathcal{S}(\kappa^w_i, m^i_l))}, \label{eq:write-addressing}
\end{align}
where $\mathcal{S}$ is a similarity function that measures the relation between the write key and memory slot:
\begin{equation}
    \mathcal{S}(x, y) = \frac{x \cdot y}{|x| |y|}.
\end{equation}
In this way, reviews with similar semantic meanings tend to have similar write weights for each memory slot.

Next, we prepare the content which should be written into the memory, and we use \emph{write content vector} $v_i \in \mathbb{R}^H$ to store this content:
\begin{equation}
    v_i = \text{Dense}(s_{i}^r),
\end{equation}
where $v_i \in \mathbb{R}^H$ represents the information of current input $s_{i}^r$ that should be written into memory.
Then, an \emph{erase vector} $E_i \in \mathbb{R}^H$ is produced to decide which dimension of the memory is useless and should be erased, as shown in Equation~\ref{eq:erase}:
\begin{equation}
    E_i = \sigma(\text{Dense}(s_{i}^r)), \label{eq:erase}
\end{equation}
where $\sigma$ denotes the sigmoid function.

To update the memory slot, we should first write the write content into the slot controlled by the write weight and then erase the useless information from the memory slot.
By combining write weight $\pi^w_{i,j} \in \mathbb{R}$, write content $v_i \in \mathbb{R}^H$, and erase vector $E_i \in \mathbb{R}^H$, we update a memory slot $m^i_j \in \mathbb{R}^H$ as follows:
\begin{equation}
    m^i_j = m^{i-1}_j \times (\mathbf{1} - \pi^w_{i,j} E_i^\intercal) + \pi^w_{i,j} v_i^\intercal ,
\end{equation}
where $\mathbf{1} \in \mathbb{R}^H$ is an $H$ matrix of ones.
After $N$ memory writing steps, memory matrix $M_{N}$ stores all information collected from this review cluster.

\subsubsection{Reading memory}

After we write all the review information into the memory slots, we need to read the reasoning result from the memory to conduct answer generation.
Similar to the procedure of writing memory, we calculate a read key to decide the read weight on each slot.
Inspired by~\citet{Vaswani2017attention}, instead of performing one single read function, we find it beneficial to use various read keys to address different slots in memory, \ie multi-head reading mechanism.
We employ a dense layer to project the $s_{N}^r$ $T$ times to obtain the \emph{read keys} $\{\kappa^r_{1}, \dots,\kappa^r_{t}, \dots, \kappa^r_{T}\}$, and we use the $t$-th read key $\kappa^{r}_{t} \in \mathbb{R}^H$ as an example to illustrate the process:
\begin{equation}
    \kappa^r_{t} = \text{Dense}_{t}(s_{N}^r). \label{eq:read-key-gen}
\end{equation}
On each of these read keys, we apply the similarity function $\mathcal{S}$, yielding $t$-th \emph{read weight} $\pi^r_{t,j} \in \mathbb{R}$ for $j$-th memory slot, shown in Equation~\ref{eq:read-addressing}:
\begin{align}
    \pi^r_{t,j} &= \frac{\exp(\mathcal{S}(\kappa^r_{t}, m^N_j))}{\sum^S_{l=1} \exp(\mathcal{S}(\kappa^r_{t}, m^N_l))} . \label{eq:read-addressing}
\end{align}
In this way, multi-head addressing allows the model to address suitable read location from different read key representation subspaces in different positions.
Finally, these read weights $\pi^r_{\cdot}$ are used to produce a weighted sum of memory slots, as shown in Equation~\ref{eq:read-out}:
\begin{align}
    r_{t} &= \sum^{S}_{j=1} m^N_j \pi^r_{t,j} , \label{eq:read-out}\\
    o^r &= W_\nu s_{N}^r + W_r [r_{1} \oplus \cdots \oplus r_{T}] , \label{eq:mem-out}
\end{align}
where $\oplus$ denotes the concatenation between vectors, $r_{t} \in \mathbb{R}^H$ is the readout vector of the $t$-th read head and $o^r \in \mathbb{R}^H$ is the output of this memory reasoning module for current review cluster.
Hence, the output of the reasoning module $o^r$ can be seen as the representation of reasoning result in this cluster of reviews with respect to the question.
Recall that we have aggregated the reviews into $K$ clusters, we now again add the cluster index $k$ in following sections and use notion $o^r_{k}$ to represent the reasoning result in $k$-th cluster.

\subsection{Attributes encoder}
\label{sec:attributes}
\emph{Key-Value Memory Network} (KVMN) is shown effective in structured data utilization~\cite{miller2016key,Tu2018Learning,He2017GeneratingNA}.
Inspired by this, in MPAG we employ an KVMN to infer representations of the structured knowledge, \ie product attributes.
Embedding of attribute's key is regarded as the key in KVMN and embedding of the attribute's value is used as the value.

We first calculate the relevance between each attribute key and the given question. 
Given question $X^q$, for the $i$-th attribute $a_i=(a^k_i, a^v_i) \in A$, their matching function $\lambda$ is calculated as:
\begin{equation}
    \lambda(a_i , X^q) = \frac{\exp(h^q_{T_q} W_a e(a^k_i))}{\textstyle \sum^{T_a}_{t=1} \exp(h^q_{T_q} W_a e(a^k_t))} ,
\end{equation}
where $h^q_{T_q}$ is question representation and $W_a$ is the parameter of linear transform which converts $h^q_{T_q}$ and $e(a^k_i)$ into the same space.

Then, we use these matching scores to produce a weighted sum of all attribute values since an attribute with a high matching score is more related to the question, thus should take a larger proportion in overall attribute representation:
\begin{equation}
    o^a = \sum^{T_a}_{i=1} \lambda(a_i , X^q) e(a^v_i)  ,
\end{equation}
where $o^a$ is the output of KVMN and will be used to guide the answer generation.

\subsection{Prototype reader}
\label{sec:prototype}

To tackle the ``universal answer'' problem, in this paper, we employ the prototype editing method to generate the answer by editing the prototype instead of generating answer from scratch.
As introduced in \S\ref{sec:intro}, the prototype question-answer pair is retrieved according to its similarity to the current question.
A prototype answer $\Tilde{X}^q$ and a prototype question $\Tilde{Y}$ are given to our prototype reader to assist the generation process.
Prototype reader in MPAG learns to extract the \emph{answer skeleton}, \ie template words in the prototype answer that are not highly related to the prototype question, to be reused in generating the new answer to increase the diversity of sentence pattern.
We first employ Bi-RNN to model the temporal interactions between words in prototype question $\Tilde{X}^q$ and answer $\Tilde{Y}$:
\begin{align}
\Tilde{h}^q_t &= \text{Bi-RNN}_{q}(e(\Tilde{x}^q_t), \Tilde{h}^q_{t-1}), \\
\Tilde{h}^a_t &= \text{Bi-RNN}_{a}(e(\Tilde{y}_t), \Tilde{h}^a_{t-1}),
\end{align}
where $\Tilde{h}^q_t$ and $\Tilde{h}^a_t$ denotes the hidden state of $t$-th step in Bi-RNN for $t$-th word in $\Tilde{X}^q$ and $\Tilde{Y}$ respectively. 

We then employ an attention mechanism to analyze the dependency between $\Tilde{h}^q_t$ and $\Tilde{h}^a_t$ to learn answer skeleton.
Then the dependency will be used to extract the answer skeleton from prototype answer which is not highly related to the prototype question.
The attention is derived from a shared similarity matrix $D \in \mathbb{R}^{T_q \times T_y}$, which is calculated by each word of prototype question $\Tilde{X}^q$ and prototype answer $\Tilde{Y}$.
$D_{ij}$ here indicates the similarity between the $i$-th question word $\Tilde{x}^q_i$ in the question and the $j$-th answer word $\Tilde{y}_j$ in the answer and is computed as:
\begin{equation}
\begin{aligned}
    D_{ij} &= \alpha(\Tilde{h}^q_i, \Tilde{h}^a_j), \\
    \alpha(x, y) &= w^\intercal [x \oplus y \oplus (x \times y)] ,
\end{aligned}
\label{eq:alpha}
\end{equation}
where $\alpha$ is a trainable scalar function that encodes the similarity between two input vectors.
We let $d_t = \text{mean}(D_{:t}) \in \mathbb{R}$ represent the attention weight on the $t$-th prototype answer word by prototype question words, and multiply with the corresponding prototype answer hidden state $\Tilde{h}^a_t$, resulting in an answer skeleton, $\hat{h}^a_t$.
In this way, the module assigns high importance weights to the words which can be reused in a new answer.

\subsection{Answer editor}
\label{sec:editor}

\begin{figure}
    \centering
    \includegraphics[scale=0.55]{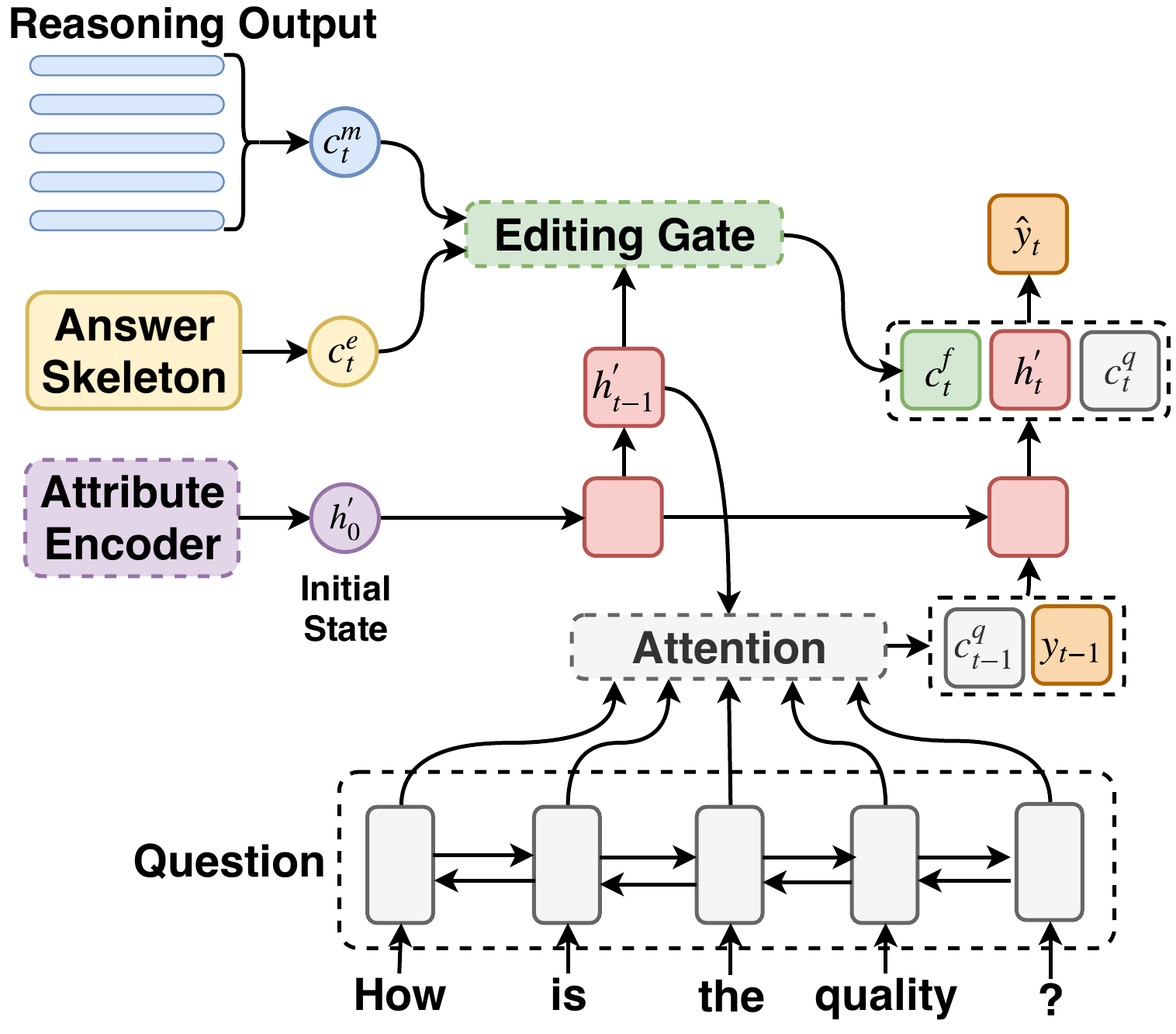}
    \caption{
    An overview of the answer editor. In this module, we incorporate reasoning result, answer skeleton and product attribute into generating the answer.
    }
    \label{fig:decoder}
\end{figure}

To generate a diverse and meaningful answer, we propose an RNN-based decoder which incorporates outputs of reasoning result, answer skeleton, question, and attributes.
Figure~\ref{fig:decoder} illustrate the framework of the answer decoder.
To force the decoder focus on the current question and product, we first apply a linear transform on the concatenation of question vector representation $h^q_{T_q}$ and attributes $o^a$, and then use this vector $h^{'}_0$ as the initial state of the $\text{LSTM}$ shown in Equation~\ref{eq:init-state}.
The $t$-th decoding step is shown in Equation~\ref{eq:dec-step}:
\begin{align}
    h^{'}_0 &= W_e \left[ h^q_{T_q} \oplus o^a \right] + b_e , \label{eq:init-state} \\
    h^{'}_t &= \text{LSTM} \left(h^{'}_{t-1}, [c^q_{t-1} \oplus e(y_{t-1})]\right) ,  \label{eq:dec-step}
\end{align}
where $W_e, b_e$ are trainable parameters, $h^{'}_t$ is the hidden state of $t$-th decoding step, $c^q_{t-1}$ is the context vector produced by the standard attention mechanism~\cite{Bahdanau2015Neural} over the hidden states $h^q_{\cdot}$ of question.
To dynamically fuse the answer skeleton and reasoning result into the generation process, we come up with an editing context vector $c^e_t$ and a memory context vector $c_{t}^{m}$.

Editing context vector $c^e_t$ is used to dynamically collect useful prototye information from answer skeleton according to current decoding state.
We first show how editing context vector $c^e_t$ is calculated from answer skeleton, as shown in Equation~\ref{eq:editing-context-align-score}-\ref{eq:edit-vector}:
\begin{align}
    \delta_{it} &= \frac{\exp(f(\hat{h}^a_i, h^{'}_t))}{\sum_{j}^{T_y} \exp(f(\hat{h}^a_j, h^{'}_t))} , \label{eq:editing-context-align-score}\\
    f(\hat{h}^a_i, h^{'}_t) &= h^{'}_t W_f \hat{h}^a_i , \label{eq:context-matching}\\
    c^e_t &= \sum_{j}^{T_y} \delta_{jt} \hat{h}^a_j , \label{eq:edit-vector}
\end{align}
where $\hat{h}^a_i$ is the $i$-th hidden state in the answer skeleton, $f$ is a bi-linear matching function which models the relationship between current decoding state $h^{'}_t$ and each hidden state $\hat{h}^a_i$ of answer skeleton.

As for memory context vector, remember in \S\ref{sec:memory}, there are $K$ reasoning results for each cluster $\{o^r_{1},\dots,o^r_{K}\}$.
We thus employ a dynamic fused method to produce a context vector of all the reasoning results:
\begin{align}
    \epsilon_{kt} &= \frac{\exp(f(o^r_{k}, h^{'}_t))}{\sum_{j}^{K} \exp(f(o^r_{j}, h^{'}_t))} ,  \\
    c^m_t &= \sum_{j}^{K} \epsilon_{jt} o^r_{j} ,  \label{eq:mem-context-vector}
\end{align}
where $f$ is the same function used in Equation~\ref{eq:context-matching} with different trainable parameters.

In each decoding step, we use an \emph{editing gate} to decide which information should be used in generating current word between prototype words and reasoning result. 
And editing gate is used as an threshold to adjust the proportion of editing context vector and memory context vector.
Now we show how to combine edit context vector $c^e_t$ with memory context vector $c^m_t$ by an \emph{editing gate} $\gamma_t$.
$\gamma_t \in \mathbb{R}$ is determined by decoder state $h^{'}_t$ and is used to decide the importance of edit and memory context vectors at $t$-th decoding step, shown in Equation~\ref{eq:editing-gate}.
And then we mix the editing context vector and memory context vector together using editing gate as shown in Equation~\ref{eq:mix-context}:
\begin{align}
    \gamma_t& = \sigma \left( Dense(h^{'}_t) \right)  \label{eq:editing-gate} , \\
    c^f_t &= \left[ \gamma_t c^m_t \oplus \left( 1 - \gamma_t \right) c^e_t \right] , \label{eq:mix-context}
\end{align}
where $\sigma$ denotes the sigmoid function, and $c^f_t$ dynamically mixes the information of memory and edit context vector.

Intuitively, to generate the answer word, there are three parts of information sources should be incorporated into generation process: prototype and reasoning result, current question and decoding state.
Finally, we concatenate $c^f_t$ with question context vector $c_{t}^{q}$ and decoder hidden state $h^{'}_t$ and apply a fully-connection layer on these vectors.
Then, we predict the output word distribution $P_{v}$ over all the words:
\begin{align}
    h^o_t &= W_o [h^{'}_t \oplus c^f_t \oplus c^q_t] + b_o , \label{eq:output-proj}\\
    P_{v} &= \text{softmax} \left(W_v h^o_t + b_v \right).
\end{align}
where $W_o, W_v, b_o, b_v$ are all trainable parameters.
Our objective function is the negative log likelihood of the target word $y_t$, shown in Equation~\ref{eq:loss}:
\begin{equation}
    \mathcal{L} = - \sum^{T_y}_{t=1} \log P_{v}(y_t) . \label{eq:loss}
\end{equation}
We employ the gradient descent method to update all parameters to minimize this loss function. 
\section{Experimental Setup}
\label{section5}

\subsection{Research Questions}\label{sec:research-question}

We list four research questions that guide the remainder of the paper:
\begin{enumerate}
  \item \textbf{RQ1} (See \S\ref{sec:overall-exp}): What is the overall performance of MPAG? Does it outperform state-of-the-art baselines?
  \item \textbf{RQ2} (See \S\ref{sec:cluster-exp}): What is the effect of the review clustering in MPAG?
  \item \textbf{RQ3} (See \S\ref{sec:reasoning-exp}): 
  Does the saliency score (calculated in Equation~\ref{eq:salience-score}) explain why the generated answer holds the corresponding opinion?
  \item \textbf{RQ4} (See \S\ref{sec:editor-exp}): Can the answer editor in MPAG learn a useful answer skeleton?
\end{enumerate}

\subsection{Dataset}

We conduct experiments on a large-scale real-world product aware question-answering dataset proposed by~\citet{Gao2019Product}.
This dataset is collected from JD.com, one of the largest e-commerce websites in China. 
On this website, users can post a question about the product.
Most questions are asking for experience of the user who has already bought the product.
This dataset is available at \url{https://github.com/gsh199449/productqa}.
It includes question-answering pairs, a large number of reviews, and product attributes.
Most questions in the dataset are about personal user experience.
In this paper, The only difference from~\cite{Gao2019Product} is that, rather than using BM25 to select a small number of review for each question, we retrieve up to 100 relevant reviews to obtain more information.
We also follow the retrieval method proposed by~\citet{Wu2018ResponseGB} to collect a prototype question-answer pair for each question.
We remove all QA pairs without any relevant review and split the whole dataset into training and testing sets.
In total, our dataset contains cover 469,953 products from 38 categories.
We use all the training dataset as our retrieve database.
The average review and attribute numbers of a product are 59.1 and 9.0, respectively.
The average lengths of a question and ground truth answer are 9.03 and 10.3 words, respectively.

\subsection{Evaluation Metrics}

To evaluate the methods, we employ BLEU~\cite{Papineni2002BleuAM} to measure the lexical unit overlapping (\eg unigram, bigram) between the generated answer and ground truth.
Following~\cite{Serban2017AHL, Xu2017NeuralRG, Tao2018Get, Gao2019Product}, we also use embedding-based metrics~\cite{forgues2014bootstrapping} (including Embedding Average, Embedding Greedy and Embedding Extreme) to compute their semantic similarity.
Besides, to quantitatively evaluate the safe answer problem, we use the distinct metric~\cite{Li2016ADO}, which evaluates the diversity of the generated answers by calculating the number of distinct unigrams and bigrams.

Since only using automatic evaluation metrics can be misleading~\cite{Stent2005EvaluatingEM}, we also conduct human evaluation. 
Three annotators are invited to judge the quality of 100 randomly sampled answers generated by different models. 
These annotators are all well-educated Ph.D. students and are all native speakers. 
Two of them have a background of NLP while another annotator does not major in computer science.
The statistical significance of two runs is tested using a two-tailed paired t-test and is denoted using \dubbelop (or \dubbelneer) for strong significance for $\alpha = 0.01$. 

\subsection{Comparisons} \label{sec:baselines}

\begin{table}[t]
\centering
\caption{Ablation models for comparison.}
\label{tab:ablations}
\begin{tabular}{@{}l@{~}l}
\toprule
Acronym & Glossary \\
\midrule
MPAG-P &  \multicolumn{1}{p{6cm}}{ w/o \textbf{P}rototype answer}\\
MPAG-M &  \multicolumn{1}{p{6cm}}{ w/o \textbf{M}emory module}\\
MPAG-K &  \multicolumn{1}{p{6cm}}{ w/o \textbf{K}-means Clustering}\\
DNC    &  \multicolumn{1}{p{6cm}}{ Replace Review Reasoning with DNC}\\
SDNC   &  \multicolumn{1}{p{6cm}}{ Replace SRU with LSTM}\\
\bottomrule
\end{tabular}
\end{table}

To prove the effectiveness of each module, we conduct ablation studies as shown in Table~\ref{tab:ablations}.
We remove each key module in our proposed model, and then form three baseline methods \texttt{MPAG-P}, \texttt{MPAG-M}, and \texttt{MPAG-K}.
Due to the fact that our review reasoning module takes inspiration from DNC~\cite{graves2016hybrid}, we also use the original DNC network to replace our review reasoning module (shown in \S\ref{sec:memory}) as a baseline method, named as \texttt{DNC}.
To examine the effectiveness of SRU compared with LSTM cell~\cite{graves2016hybrid}, we replace SRU with the LSTM cell, named as \texttt{SDNC}.

Apart from the ablation study, we also compare our model with the following baselines: 
\begin{enumerate}
  \item \texttt{BM25} is a bag-of-words retrieval function that ranks a set of reviews based on the question terms appearing in each review.
  We use the top review of the ranking list as the answer.
  \item \texttt{TF-IDF} (Term Frequency-Inverse Document Frequency) is a numerical statistic that is intended to reflect how important a question word is to a review.
  We use this statistic to model the relevance between review and question and select the most similar review as the answer of the question.
  \item \texttt{S2SA} is the Sequence-to-Sequence (seq2seq) framework~\cite{Sutskever2014SequenceTS} which has been proposed for language generation task. 
  We use the seq2seq framework which is equipped with the attention mechanism~\cite{Bahdanau2015Neural} and copy mechanism~\cite{Gu2016IncorporatingCM} as a baseline method. 
  Attention mechanism~\cite{Bahdanau2015Neural} has been proposed to tackle the alignment between the input sequence and the generated sequence. 
  Copy mechanism~\cite{Gu2016IncorporatingCM} has been widely used in the text generation task to tackle the OOV problem, which can copy some words from the input sequence when generating new text.
  The input sequence is a question and the ground truth output sequence is the answer.
  \item \texttt{S2SAR} is a simple method which can incorporate the review information when generating the answer.
  Different from the S2SA, we use an RNN to read all the reviews and concatenate the final state of this RNN with encoder final state as the initial state of decoder RNN.
  \item \texttt{SNet}~\cite{Tan2018snet} is a two-stage state-of-the-art model which extracts some text spans from multiple documents context and synthesis the answer from those spans.
  Due to the difference between our dataset and MS-MARCO~\cite{Nguyen2016MSMA}, our dataset does not have text span label ground truth for training the evidence extraction module.
  So we use the predicted extraction probability to do weighted sum the original review word embeddings, and use this representation as extracted evidence to feed into the answer generation module.
  \item \texttt{QS} is the query-based summarization model proposed by Hasselqvist \etal~\cite{Hasselqvist2017QueryBasedAS}. 
  Accordingly, we use product reviews as an original passage and answer as a summary.
  \item \texttt{PAAG} is the product-aware answer generation model proposed in our previous work~\cite{Gao2019Product}.
  \item \texttt{Proto} is the prototype editing response generation model in dialog generation task proposed by~\cite{Wu2018ResponseGB}.
  \item \texttt{Re\textsuperscript{3}Sum} is the text summarization model~\cite{Cao2018Retrieve}, which retrieves summaries and conducts template aware summary generation.
  \item \texttt{RAGE} is a review-driven e-commerce question answering model using convolutional sequence generation~\cite{Chen2019Review}.
\end{enumerate}

\subsection{Implementation Details}

All parameters in our model are randomly initialized.
The number of K-means cluster is set to $3$. 
The maximum number of reviews in each cluster is $20$.
The RNN-based networks have $512$ hidden units and the dimension of a word embedding is $256$.
We limit the length of the question and answer sentence to $25$ words and review sentence to $30$ words.
The beam search algorithm is employed with a beam width of $4$.
We use Adagrad~\cite{Duchi2010AdaptiveSM} to update the parameters with a learning rate of $0.1$ and training batch size of $64$.
Our model is implemented via TensorFlow~\cite{abadi2016tensorflow} framework and trained on an NVIDIA GTX 1080 Ti GPU. %

\section{Experimental Result}
\label{section6}

\newcommand{\cbkgrnd}{\cellcolor{blue!15}}
\newcommand{\phantomtriangle}{\phantom{\dubbelop}}

\begin{table}[t]
\centering
\caption{Automatic evaluation comparison with baselines.}
\begin{tabular}{@{}lcc cc c@{}}
\toprule
& BLEU & BLEU1 & BLEU2 & BLEU3 & BLEU4 \\
\midrule
\multicolumn{5}{@{}l}{\emph{Text generation methods}}\\

S2SA & 1.62\phantom{0} & 15.48\phantom{0} & 3.14\phantom{0} & 0.83\phantom{0} & 0.17\phantom{0} \\
S2SAR & 1.75\phantom{0} & 15.17\phantom{0} & 3.22\phantom{0} & 0.91\phantom{0} & 0.21\phantom{0}\\
RAGE & 0.22\phantom{0} & 8.58\phantom{0} & 0.72\phantom{0} & 0.05\phantom{0} & 0.01\phantom{0} \\
SNet & 0.96\phantom{0} & 13.70\phantom{0} & 2.54\phantom{0} & 0.40\phantom{0} & 0.06\phantom{0}\\
QS & 1.68\phantom{0} & 15.50\phantom{0} & 2.95\phantom{0} & 0.83\phantom{0} & 0.21\phantom{0}\\
Proto & 2.83\phantom{0} & 21.80\phantom{0} & 5.36\phantom{0} & 1.33\phantom{0} & 0.41\phantom{0}\\
Re\textsuperscript{3}Sum & 2.83\phantom{0} & 22.03\phantom{0} & 5.62\phantom{0} & 1.50\phantom{0} & 0.34\phantom{0}\\
\cbkgrnd PAAG & \cbkgrnd 2.02\phantom{0} & \cbkgrnd 16.22\phantom{0} & \cbkgrnd 3.57\phantom{0} & \cbkgrnd 1.03\phantom{0} & \cbkgrnd 0.28\phantom{0}\\
MPAG & \textbf{3.96}\dubbelop & \textbf{24.25}\dubbelop & \textbf{6.68}\dubbelop & \textbf{2.09}\dubbelop & \textbf{0.73}\dubbelop\\
\midrule
\multicolumn{5}{@{}l}{\emph{Sentence extraction methods}}\\
BM25 & 0.41\phantom{0} & 6.96\phantom{0} & 0.71\phantom{0} & 0.13\phantom{0} & 0.04\phantom{0}\\
TF-IDF & 0.25\phantom{0} & 5.55\phantom{0} & 0.51\phantom{0} & 0.08\phantom{0} & 0.02\phantom{0}\\
\bottomrule
\end{tabular}
\label{tab:comp_auto_baselines}
\end{table}

\begin{table}[t]
  \centering
  \caption{Embedding scores comparison between baselines.}
  \begin{tabular}{@{}lcc c@{}}
  \toprule
  & Average & Greedy  & Extrema \\
  \midrule
  \multicolumn{4}{@{}l}{\emph{Text generation methods}}\\
  S2SA & 0.410013\phantom{0} & 98.653415\phantom{0} & 0.269461\phantom{0} \\
  S2SAR & 0.419979\phantom{0} & 99.742679\phantom{0}& 0.278666\phantom{0} \\
  SNet & 0.397162 & 95.791356 & 0.277781 \\
  QS & 0.400291 & 93.255031 & 0.252164 \\
  \cbkgrnd PAAG & \cbkgrnd 0.424218 & \cbkgrnd 103.912364 & \cbkgrnd 0.288321 \\
  MAPG & \textbf{0.526868}\dubbelop & \textbf{139.3584}\dubbelop & \textbf{0.432037}\dubbelop \\
  \midrule
  \multicolumn{4}{@{}l}{\emph{Sentence extraction methods}}\\
  BM25 & 0.325946 & 76.814465 & 0.172976 \\
  TF-IDF & 0.308293 & 85.020442 & 0.155390 \\
  \bottomrule
  \end{tabular}
  \label{tab:comp_emb_baselines}
  \end{table}

\subsection{Overall Performance}
\label{sec:overall-exp}

At the beginning, we address the research question \textbf{RQ1}.
In the Table~\ref{tab:comp_auto_baselines}, the significant differences are with respect to \texttt{PAAG} (row with shaded background).
In these experimental results, we see that PAAG achieves a 111\% increment over the state-of-the-art question answering baseline SNet in terms of BLEU, which demonstrates the effectiveness of using adversarial training method and incorporating product attributes.
For our newly proposed model MPAG, we can see that MPAG achieves a 93.18\% increase over the state-of-the-art baseline \texttt{PAAG} in terms of BLEU, and the improvements are all significant (with p-value < 0.05).
As for the prototype-based baselines \texttt{Proto} and \texttt{Re\textsuperscript{3}Sum}, they all outperform our previously proposed model \texttt{PAAG}.
This suggests that introducing prototype answer and employing the novel memory network can help the model to generate better answers.
Despite the prototype-based methods obtain the help from the prototype question-answer pairs, these methods can not beat our proposed MPAG, since they fail to fully utilize the prototype answer due to their lack of reasoning ability.

We also employ the embedding metric~\cite{forgues2014bootstrapping} as another automatic evaluation metric which goes beyond simple N-gram matches.
From this experiment, we can find that MAPG achieves a 24.20\% increase over the state-of-the-art baseline \texttt{PAAG} in terms of Average, and the improvements are all significant (with p-value < 0.05).
The embedding-based metric measures the semantic matches between the generated answer and ground-truth answer, and this suggests that our newly proposed model can generate answers with high semantic consistency with the ground-truth answer.

For human evaluation, annotators rate each generated answer according to two objectives:
(1) \textbf{Consistency}: Is the meaning of the answer consistent with the question?
(2) \textbf{Fluency}: Is the generated answer well-written?
The rating score ranges from 1 to 3, with 3 being the best.
The results are shown in Table~\ref{tab:comp_human_baslines}, and MPAG outperforms \texttt{PAAG} by 4.0\% and 10.3\% in terms of fluency and consistency.
The paired student t-test demonstrates the significance of the above results. 
The kappa statistics are 0.56 and 0.53 for fluency and consistency respectively, which indicates moderate agreement between annotators\footnote{\citet{landis1977measurement} characterize kappa values < 0 as no agreement, 0-0.20 as slight, 0.21-0.40 as fair, 0.41-0.60 as moderate, 0.61-0.80 as substantial, and 0.81-1 as almost perfect agreement.}.

\subsection{Effect of Clustering}
\label{sec:cluster-exp}

In this section, we address the research question \textbf{RQ2}.
To verify the effectiveness of the review clustering, we randomly aggregate the reviews into three clusters instead of using the K-Means algorithm, and feed the review clusters to the model named \texttt{MPAG-K}.
The automatic evaluation results in Table~\ref{tab:comp_auto_ablations} show that the performance of \texttt{MPAG-K} decreases by 4.75\% compared with \texttt{MPAG}, in terms of BLEU1.
This demonstrates the necessity of clustering reviews into the corresponding aspect.

\subsection{Effect of Reasoning Module}
\label{sec:reasoning-exp}

Next, we turn to the research question \textbf{RQ3}.
In addition to the ablation study in Table~\ref{tab:comp_auto_ablations}, where \texttt{MPAG-M} decreases by 9.63\% compared with \texttt{MPAG} in terms of BLEU, 
we also compare reasoning module with \texttt{DNC} and \texttt{SDNC}.
The fact that \texttt{SDNC} performs better than \texttt{DNC} demonstrates that vanilla DNC is not suitable for this scenario though it consists of complicated structures such as temporal memory linkage and dynamic memory allocation.
However, SRU further improves the BLEU score by 12.24\%, compared with \texttt{SDNC}, which further verifies its superiority.

Note that, in \S\ref{sec:memory}, the SRU-based network learns to assign high weights to reviews that contain more useful information related to the question, and thus the saliency weight makes MPAG an explainable answer generator.
Next we conduct two experiments to examine whether the saliency weight can faithfully reflect the importance of each review.

\begin{table}[t]
\centering
\caption{Human evaluation comparison with main baseline.}
\begin{tabular}{@{}lccc@{}}
\toprule
& Fluency & Consistency & Correctness\\ 
\midrule
PAAG & 2.75\phantom{0} & 2.52\phantom{0} & 69.0\%\\
MPAG & \textbf{2.86}\dubbelop & \textbf{2.78}\dubbelop & \textbf{91.0}\%\dubbelop \\
\bottomrule
\end{tabular}
\label{tab:comp_human_baslines}
\end{table}

\begin{table}[t]
\centering
\caption{Automatic evaluation comparison between ablation models.}
\begin{tabular}{@{}lcc cc c@{}}
\toprule
& BLEU & BLEU1  & BLEU2 & BLEU3 & BLEU4 \\
\midrule
MPAG-P & 2.13 & 19.47 & 4.14 & 1.03 & 0.25 \\
MPAG-M & 3.61 & 23.04 & 6.37 & 1.95 & 0.59 \\
MPAG-K & 3.54 & 23.15 & 6.35 & 1.82 & 0.58 \\
DNC    & 3.41 & 22.69 & 6.10 & 1.72 & 0.57 \\
SDNC   & 3.53 & 23.28 & 6.29 & 1.81 & 0.58 \\
\bottomrule
\end{tabular}
\label{tab:comp_auto_ablations}
\end{table}

\begin{table}[t]
\centering
\caption{Diversity evaluation comparison with baselines.}
\begin{tabular}{@{}lcccc@{}}
\toprule
& Distinct-1 & Distinct-2 & Distinct-3 & Distinct-4\\ 
\midrule
PAAG                     & 0.0310 & 0.1129 & 0.2299 & 0.3495 \\
Re\textsuperscript{3}Sum & 0.0291 & 0.1299 & 0.2826 & 0.4429 \\
Proto                    & 0.0273 & 0.1364 & 0.2946 & 0.4535 \\
RAGE                     & 0.0377 & 0.0476 & 0.1945 & 0.4439 \\
MPAG                     & \textbf{0.0392} & \textbf{0.1902} & \textbf{0.3959} & \textbf{0.5763} \\
\bottomrule
\end{tabular}
\label{tab:comp_distinct}
\end{table}

\subsubsection{Visualization of Salience Weight}
\label{sec:vis-salience}

\begin{figure}[!t]
  \subfigure{ 
    \includegraphics[clip,width=0.3\columnwidth]{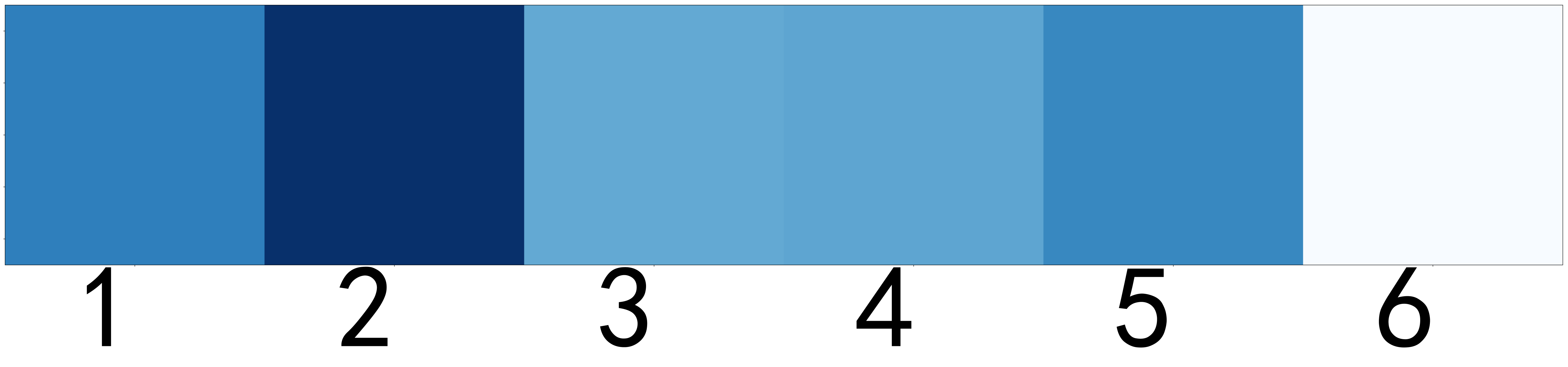}
    }
  \subfigure{
    \includegraphics[clip,width=0.45\columnwidth]{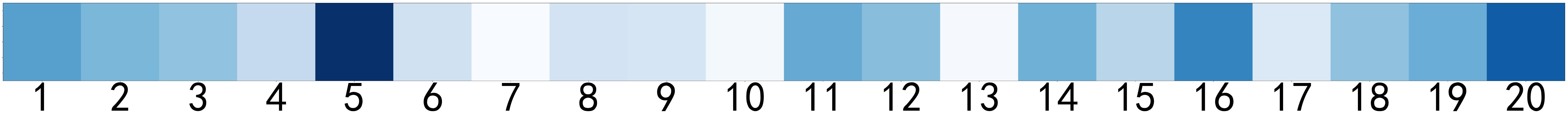}}
    \caption{Visualizations of salience weights over several reviews. The number is the review index and the review contents are listed in Section~\ref{sec:vis-salience}.}
  \label{fig:sru-case}
\end{figure}

We first visualize the saliency weights of a review cluster in Figure~\ref{fig:sru-case} and determine whether it is in accordance with human intuition.
The darker the color, the more important the corresponding review.
For the top figure in Figure~\ref{fig:sru-case}, the question is ``\begin{CJK*}{UTF8}{gbsn}鱼竿质量怎么样\end{CJK*}'' (What is the quality of this fishing rod?).
The second review is ``\begin{CJK*}{UTF8}{gbsn}质量很好，手感不错，轻巧\end{CJK*}'' (The quality of the fishing rod is very good, feeling good, lightweight) and has the highest weight, while the sixth review is ``\begin{CJK*}{UTF8}{gbsn}物品已收到，买一送一，手竿比较轻，比较软，等有时间再去用用看质量如何\end{CJK*}'' (Product received, buy one get one free. The handcuffs are light and soft. I will examine the quality when I have time to fish), which has the lowest weight.
This is consistent with human intuition.

In the bottom of Figure~\ref{fig:sru-case}, the question is ``\begin{CJK*}{UTF8}{gbsn}大家觉得这款剃的干净吗？尤其是死角的地方\end{CJK*}'' (Do you think this razor can shave cleanly? Especially the corners of the face).
We can see that the fifth review has a higher weight than the seventh review.
The fifth review is ``\begin{CJK*}{UTF8}{gbsn}送给老公的礼物，非常好，剃得很干净\end{CJK*}'' (I use it as the gift for husband, very good, shaved very cleanly) and the seventh review is ``\begin{CJK*}{UTF8}{gbsn}剃须刀很漂亮，用的也很顺手，剃须功能棒棒的，很喜欢这款\end{CJK*}'' (The razor is very beautiful, easy to use, it is great, I like it very much).
We can easily see that, the fifth review is more useful than the seventh with respect to the question.

\begin{figure}[!t]
  \centering
    \includegraphics[clip,width=0.5\columnwidth]{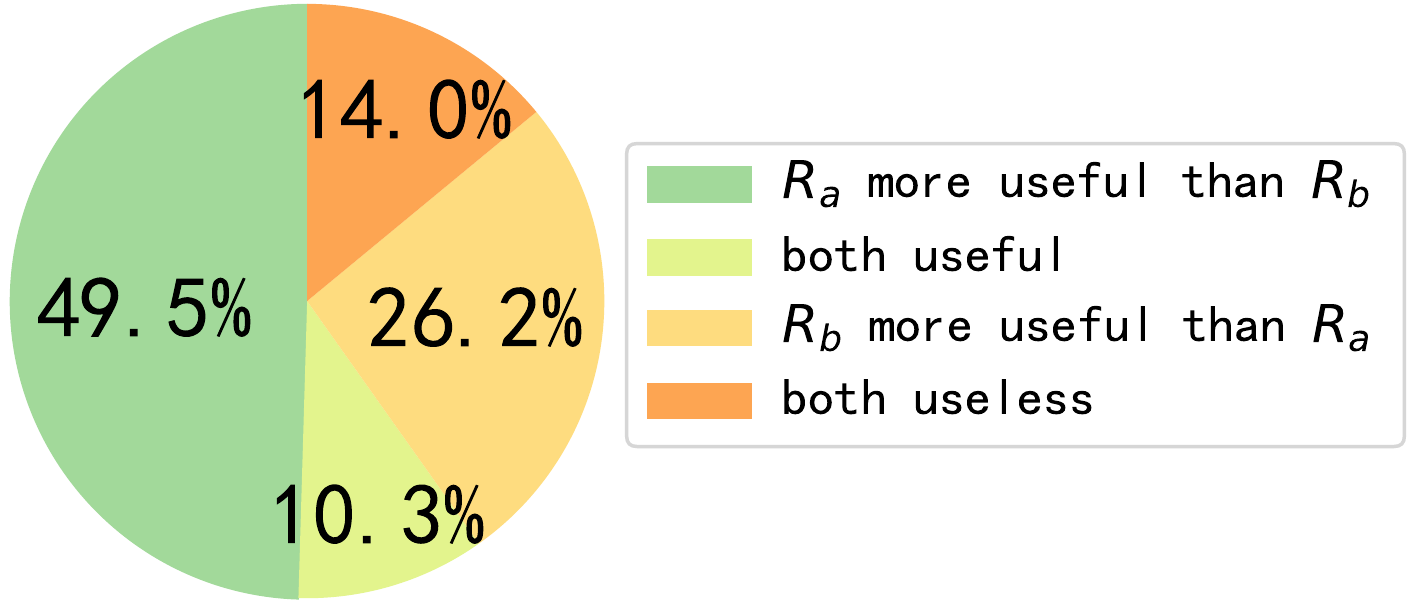}
    \caption{Pairwise explanation annotation result.}
  \label{fig:pairwise-explanation}
\end{figure}

\subsubsection{Quantitative Usefulness Evaluation}
\label{sec:pairwise-exp}

It is intuitive to conduct a pairwise evaluation on whether our high weight reviews are as helpful as the rated useful ones selected by metrics based on word similarity such as \texttt{BM25}.
Hence, we randomly select 100 data samples from two groups of reviews for comparison.
Specifically, we choose the review with the highest weight score selected by \texttt{MPAG} ($R_a$) and \texttt{BM25} ($R_b$), along with the question.
The order of these two reviews is randomly shuffled and four choices are listed to annotators:
(1) $R_a$ is more useful than $R_b$;
(2) $R_b$ is more useful than $R_a$;
(3) $R_a$ and $R_b$ are almost the same, both useful;
(4) $R_a$ and $R_b$ are almost the same, both useless.

The pairwise evaluation results are shown in Figure~\ref{fig:pairwise-explanation}.
For 49.5\% of data samples, $R_a$ is more useful than $R_b$.
For 10.3\% of data samples, the annotators think both reviews are useful and find it hard to judge which is better.
Finally, only 14\% of data samples, the given explanation is useless.
Therefore, we conclude that, in most cases, MPAG achieves good performance in providing explanations.

\subsection{Effect of Answer Editor}
\label{sec:editor-exp}

\begin{figure}[!t]
  \centering
    \includegraphics[clip,width=0.3\columnwidth]{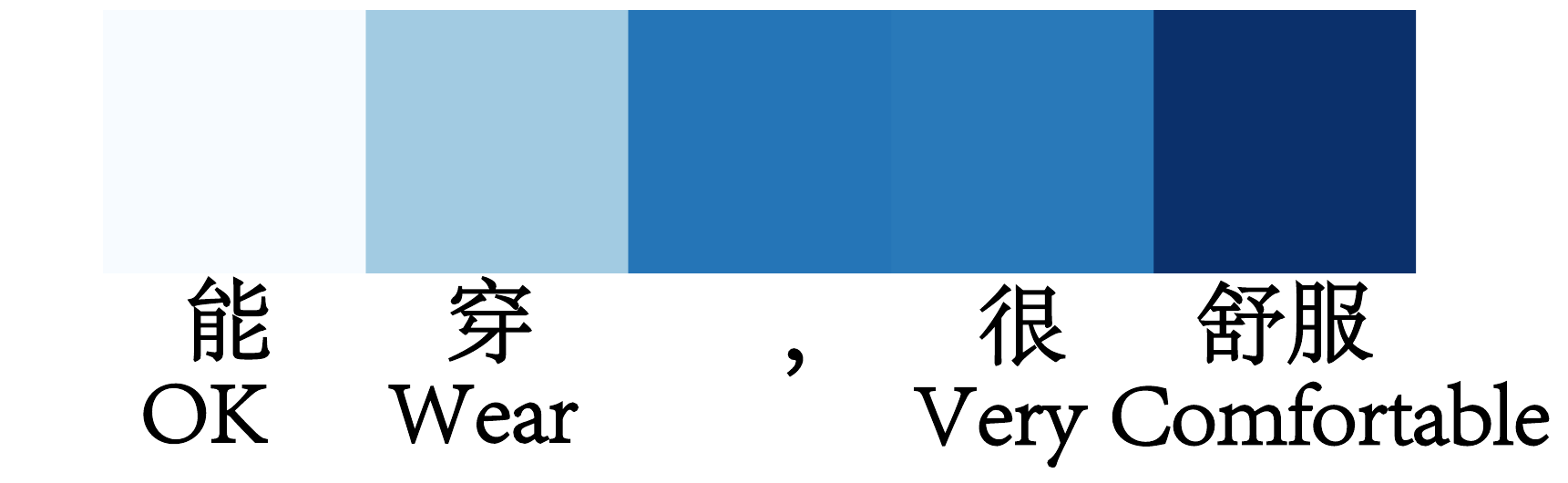}
    \caption{Visualization of editing gates. The darker the color is, the more information of this word comes from reasoning context vector.}
  \label{fig:editing-gate-case}
\end{figure}

Lastly, we address the research question \textbf{RQ4}.
In Table~\ref{tab:comp_auto_ablations}, compared with \texttt{MPAG}, the performance of \texttt{MPAG-P} drops most by 24.61\% in terms of BLEU1.
This observation suggests that prototype question-answer pairs are helpful, and our model successfully learns how to utilize them.
To examine whether the prototype method can alleviate the safe answer problem, we first evaluate by the distinct metric as shown in Table~\ref{tab:comp_distinct}. 
\texttt{MPAG} outperforms the main baseline \texttt{PAAG} by 26.45\% and 68.46\% in terms of Distinct-1 and Distinct-2, respectively.
Furthermore, human evaluation is also conducted to evaluate the proportion of safe answers in generation results. 
About 25\% of the \texttt{PAAG} outputs are the safe answers, while our model produces only 20\% safe answers, a great decrease.
Nearly 95\% of answers generated by our model are distinct, which makes our model far more practical than PAAG. 
Moreover, the corresponding kappa score for the inter-annotator agreement is $0.37$.
The above experiments demonstrate MPAG is indeed helpful for alleviating the safe answer problem.

Since the prototype context improves the performance of \texttt{MPAG} by such a large margin, it is possible that the prototype answer is already a good answer to the question, and \texttt{MPAG} directly copies the prototype answer as output.
To examine whether the improvement is brought by the prototype answer or the prototype editing module, we calculate the BLEU scores between the prototype answer and the ground truth answer, and obtain 2.74, 18.63, 4.34, 1.21, 0.58 in terms of BLEU and BLEU1 to BLEU4, respectively.
In contrast, the scores of MPAG are 3.96, 24.25, 6.68, 2.09, 0.73, \ie higher than the scores obtained by the prototype answer, in all metrics.
This means that the original prototype answer is not good enough and our answer editor module has an efficient revision ability.

To further investigate the editing module, we visualize the editing gates $\gamma_t$ in Equation~\ref{eq:mix-context} and randomly pick one case, as shown in Figure~\ref{fig:editing-gate-case}.
The question for this sample is ``\begin{CJK*}{UTF8}{gbsn}我，一米六五身材，体重140斤，请问我能穿吗？\end{CJK*}'' (My height is 1.65 meters, and my weight is 150 kg. Can I wear it?) and the prototype answer is ``\begin{CJK*}{UTF8}{gbsn}能\end{CJK*}'' (It is ok).
The generated answer is ``\begin{CJK*}{UTF8}{gbsn}能穿，很舒服\end{CJK*}'' (It is ok to wear, very comfortable).
In Figure~\ref{fig:editing-gate-case}, we can see that the last word ``\begin{CJK*}{UTF8}{gbsn}舒服\end{CJK*}'' (comfortable) has the highest editing weight, which is consistent with the fact that comfortable is a reasoning result from reviews.
In contrast, the first word ``\begin{CJK*}{UTF8}{gbsn}能\end{CJK*}'' (ok) has a low editing weight, because it can be directly copied from the prototype answer.

\subsection{Case Study}

We also show a case study in Table~\ref{tab:case}, which includes a question, representative reviews, prototype question-answer pair, and answers generated by different models.
MPAG adapts the prototype answer to the new context and generate the answer, which is a correct answer proved by user experience.
In contrast, \texttt{PAAG} gives the answer ``no'' and then generates ``I put boiling water in it'', that makes up an inconsistent sentence confusing the readers.
We can see that the generated answer of our \texttt{MPAG} can effectively generate reasonable and fluent answers.

\begin{CJK*}{UTF8}{gbsn}
    \begin{table}[t]
    \centering
    \caption{Examples of the context and answers.}
    \resizebox{0.8\columnwidth}{!}{
    \begin{tabular}{l|l}
    \toprule
    \multicolumn{1}{l|}{\multirow{6}{*}{Reviews}} & \multicolumn{1}{p{0.8\columnwidth}}{很实用，可以倒开水 ，买了俩 \par (Very practical, can be filled with boiled water, I bought two)}     \\ \cline{2-2} 
    \multicolumn{1}{c|}{}                         & \multicolumn{1}{p{0.8\columnwidth}}{不漏水，用开水泡茶一样不漏 \par (This cup does not leak water, and it does not leak either when you use boiling water to make tea.)}    \\ \cline{2-2} 
    \multicolumn{1}{c|}{}                         & \multicolumn{1}{p{0.8\columnwidth}}{装开水两次第二次杯底破了！\par (I used this cup to fill the water twice, and the bottom of the cup broke when I opened the water for the second time!)}      \\ \cline{2-2} 
    \multicolumn{1}{c|}{}                         & \multicolumn{1}{p{0.8\columnwidth}}{杯子很漂亮，打开水也不烫手 \par (The cup is very beautiful, and using this cup to put the boiling water will not burn your hands.)}      \\ \cline{2-2} 
    \multicolumn{1}{c|}{}                         & \multicolumn{1}{p{0.8\columnwidth}}{玻璃很薄，买来后用开水烫了一下，就放起来了 \par (This glass is very thin. I bought it and washed it with boiling water. Then I put it away.)}      \\ \hline  
    Attributes                                    & \multicolumn{1}{p{0.8\columnwidth}}{
    功能:带杯套,
    花色: 无色 透明,
    国产 / 进口:国产,
    形状: 圆形,
    数量: 1 个,
    分类: 玻璃杯,
    容量: 301-400ml
    (
    Function: glass cup with a cover,
    Color: colorless, transparent,
    Domestic / Import: Domestic,
    Shape: round,
    Quantity: 1,
    Category: Glass,
    Capacity: 301-400ml
    ) }     \\ \hline
    Prototype question              & \multicolumn{1}{p{0.8\columnwidth}}{可不可以放猪血 \par (Can it be used to put pig blood?)} \\ \hline
    Prototype answer                & \multicolumn{1}{p{0.8\columnwidth}}{可以, 当然可以 \par (Yes, of course.)} \\ \hline
    Question                        & \multicolumn{1}{p{0.8\columnwidth}}{可以 用 开水 泡茶 吗 \par (Can I use this cup to put boiling water and make tea?)} \\
    \hline
    Reference                       & \multicolumn{1}{p{0.8\columnwidth}}{可以 \par (Yes, you can.)}   \\ \hline
    PAAG                            & \multicolumn{1}{p{0.8\columnwidth}}{不 可以 ， 我 用 的是 开水 \par (No, I put boiling water in it.)}  \\ \hline
    MPAG                            & \multicolumn{1}{p{0.8\columnwidth}}{当然 可以 ， 我 就 是 盛开水 \par (Yes, of course, I use this cup to put boiling water.)} \\ 
    \bottomrule
    \end{tabular}
    }
    \label{tab:case}
    \end{table}
    \end{CJK*}

\section{Conclusion}
\label{section7}
In our previous work, we propose the task of generating answers of product-related questions using custom reviews and product attributes, and we also propose an adversarial learning method which employs an attention-based review reader to extract question-aware facts from reviews and attributes and finally generate an answer.
Although the Wasserstein distance-based adversarial learning method is used to training the model which can reduce the probability of generating meaningless answers, the model still tends to generate safe answers.
And the model lacks of reasoning ability when extracts facts from custom reviews, which is necessary for generating accurate answer.

Motivated by these observations, in this paper, we proposed the \emph{Meaningful Product Answering Generator} (MPAG), which aims to generate a meaningful and diverse answer based on product attributes and reviews. 
Specifically, we employed a clustering algorithm to aggregate the reviews into several clusters, then we used a selective reading mechanism and read-write memory to encode these reviews so as to reason among them. 
We also used a key-value memory network to encode the product attributes.
To alleviate the safe answer problem, we incorporated a prototype question-answer pair to extract answer skeletons. 
Finally, we combined all the intermediate results into an RNN-based decoder to generate the answer.
Extensive experiments on a large-scale, real-world dataset showed that MPAG outperforms the state-of-the-art baselines and verified the effectiveness of each module in MPAG.
Besides, pairwise experiments demonstrated that MPAG is able to provide a reasonable explanation why the generated answer holds such an opinion.

In future work, we are looking forward to introducing user profile features to the model to provide personalized services.

\section*{Acknowledgments}
We would like to thank the anonymous reviewers for their constructive comments. 
We would also like to thank Anna Hennig in Inception Institute of Artificial Intelligence for her help on this paper. 
This work was supported by the National Key R\&D Program of China (2020AAA0105200) and the National Science Foundation of China (NSFC No. 61876196).
This work was supported by the Beijing Outstanding Young Scientist Program (NO. BJJWZYJH012019100020098).
Rui Yan is supported as a Young Fellow of Beijing Institute of Artificial Intelligence (BAAI).
 \clearpage
\bibliographystyle{ACM-Reference-Format}
\bibliography{sample-bibliography}

\end{document}